\documentclass{article} % For LaTeX2e
\usepackage{iclr2022_conference,times}

\usepackage{amsmath,amsfonts,bm}

\def\eqref#1{equation~\ref{#1}}
\def\1{\bm{1}}

\DeclareMathAlphabet{\mathsfit}{\encodingdefault}{\sfdefault}{m}{sl}
\SetMathAlphabet{\mathsfit}{bold}{\encodingdefault}{\sfdefault}{bx}{n}

\usepackage{hyperref}
\usepackage{url}

\usepackage{amsmath}
\usepackage{amssymb}
\newcommand{\cmark}{\ding{51}}%
\newcommand{\xmark}{\ding{55}}%
\usepackage{pifont}
\usepackage{adjustbox}
\usepackage{hyperref}
\usepackage{subcaption}
\usepackage{caption} 
\usepackage{fixltx2e}
\usepackage{multirow}
\usepackage{stfloats}

\usepackage{xurl}
\title{ME-GCN: Multi-dimensional Edge-Embedded Graph Convolutional Networks for Semi-supervised Text Classification}

\author{Kunze Wang,\textsuperscript{\rm 1*}
Soyeon Caren Han,\textsuperscript{\rm 2*$\dagger$} 
Siqu Long \textsuperscript{\rm 1}
and Josiah Poon\textsuperscript{\rm 2}  \\
School of Computer Science\\
The University of Sydney\\
NSW, Australia \\
\texttt{\textsuperscript{\rm 1}\{kwan4418, slon6753\}@uni.sydney.edu.au}\\
\texttt{\textsuperscript{\rm 2}\{caren.han, josiah.poon\}@sydney.edu.au}
}

\iclrfinalcopy % Uncomment for camera-ready version, but NOT for submission.
\begin{document}

\maketitle
\let\thefootnote\relax\footnotetext{* Equal contribution}
\let\thefootnote\relax\footnotetext{$\dagger$ Corresponding author (Caren.Han@sydney.edu.au)}

\begin{abstract}
Compared to sequential learning models, graph-based neural networks exhibit excellent ability in capturing global information and have been used for semi-supervised learning tasks. Most Graph Convolutional Networks are designed with the single-dimensional edge feature and failed to utilise the rich edge information about graphs. This paper introduces the ME-GCN (Multi-dimensional Edge-enhanced Graph Convolutional Networks) for semi-supervised text classification. A text graph for an entire corpus is firstly constructed to describe the undirected and multi-dimensional relationship of word-to-word, document-document, and word-to-document. The graph is initialised with corpus-trained multi-dimensional word and document node representation, and the relations are represented according to the distance of those words/documents nodes. Then, the generated graph is trained with ME-GCN, which considers the edge features as multi-stream signals, and each stream performs a separate graph convolutional operation. Our ME-GCN can integrate a rich source of graph edge information of the entire text corpus. The results have demonstrated that our proposed model has significantly outperformed the state-of-the-art methods across eight benchmark datasets. The code is available on: \url{https://github.com/usydnlp/ME\_GCN}
\end{abstract}

\section{Introduction}
Deep Learning models have performed well and have been widely used for text classification; however, the performance is not always satisfactory when utilising small labelled datasets. In many practical scenarios, the labelled dataset is very scarce as human labelling is time-consuming and may require domain knowledge. There is a pressing need for studying semi-supervised text classification with a relatively small number of labelled training data in deep learning paradigm. For the successful semi-supervised text classification, it is crucial to maximize effective utilization of structural and feature information of unlabelled data. 

Graph Neural Networks have recently received lots of attention as it can analyse rich relational structure, prioritize global features exploitation, and preserve global structure of a graph in embeddings. Due to these benefit, there have been successful attempts to revisit semi-supervised learning with Graph Convolutional Networks (GCN) \citep{KipfW17}. TextGCN \citep{yao2019graph} initialises the whole text corpus as a document-word graph and applies GCN. It shows potential of GCN-based semi-supervised text classification. \citet{linmei2019heterogeneous} worked on semi-supervised short text classification using GCN with topic-entity, and  \citet{liu2020tensor} proposed tensorGCN with semantic, syntactic, and sequential information. One major problem in those existing GCN-based text classification models is that edge features are restricted to be one-dimensional, which are the indication about whether there is edge or not (e.g. binary connectedness) or often one-dimensional real-value representing similarities (e.g. pmi, tf-idf). Instead of being a binary indicator variable or a single-dimensional value, edge features can possess rich information and fully incorporated by using multi-dimensional vectors. Addressing this problem is likely to benefit several graph-based classification problems but is particularly important for the text classification task. This is because the relationship between words and documents can be better represented in a multi-dimensional vector space rather than a single value. For example, word-based vector space models embed the words in a vector space where similarly defined words are mapped near to each other. Rather than using the lexical-based syntactic parsers or additional resources, words that share semantic or syntactic relationships will be represented by vectors of similar magnitude and be mapped in close proximity to each other in the word embedding. Using this multi-dimensional word embedding as node and edge features, it would be more effective to analyse rich relational information and explore global structure of a graph. Then, what would be the best way to exploit edge features in a text graph convolutional network?
According to the recently reported articles \citep{gong2019exploiting,khan2019multi,MRGCN,liu2020tensor,schlichtkrull2018modeling}, more rich information should be considered in the relations in the graph neural networks. 

In this paper, we propose a new multi-dimensional edge enhanced text graph convolutional networks (ME-GCN), which is suitable for the semi-supervised text classification. Note that the focus of our semi-supervised text classification task is on small proportion of labelled text documents with no other resource, i.e. no pre-trained word embedding or language model, syntactic tagger or parser. We construct a single large textual graph from an entire corpus, which contains words and documents as nodes. The graph describes the undirected and multi-dimensional relationship of word-to-word, document-document, and word-to-document. Each word and document are initialised with corpus-trained multi-dimensional word and document embedding, and the relations are represented based on the semantic distance of those representations. Then, the generated graph is trained with ME-GCN, which considers edge features as multi-stream signals, and each stream performs a separate graph convolutional operation. We conduct experiments on several semi-supervised text classification datasets. Our model can achieve strong text classification performance with a small proportion of labelled documents with no additional resources. The main contributions are: 
\begin{itemize}
    \item To the best of our knowledge, this is the first attempt to apply multi-dimensional edge features on GNN for text classification.
    \item ME-GCN is proposed to use corpus-trained multi-dimensional word and document-based edge features for the semi-supervised text classification.
    \item Experiments are conducted on several semi-supervised text classification datasets to illustrate the effectiveness of ME-GCN.
\end{itemize}

\section{Related Works}
\textbf{Semi-supervised text classification: } Due to the high cost of human labelling and the scarcity of fully-labelled data, semi-supervised models have received attention in text classification. Latent variable models \citep{chen2015dataless} apply topic models by user-oriented seed information and infer the documents’ labels based on category-topic assignment. The embedding-based model \citep{tang2015pte, meng2018weakly} utilise seed information to derive text (word or document) embeddings for documents and labels for text classification. \citet{yang2017improved} leveraged sequence-to-sequence Variational AutoEncoders (VAEs), and \citet{miyato2017} utilized adversarial training to the text domain by applying perturbations to the word embeddings. Graph convolutional networks (GCN) have been popular in semi-supervised learning as it shows superior global structure understanding ability \citep{KipfW17}.

\textbf{GNN for Text Classification: } Graph Neural Networks have successfully used in various NLP tasks \citep{bastings2017graph, tu2019multi, cao2019multi, mrgnn}. 
\citet{yao2019graph} proposed the Text Graph Convolutional Networks by applying a basic GCN \citep{KipfW17} to the text classification task. In their work, a text graph for the whole corpus is constructed; word and document nodes are initialised with one-hot representation and edge features are represented as one-dimensional real values, such as PMI, TF-IDF.  Several studies have attempted multiple different graph alignments using knowledge graph or semantic/syntactic graph. \citet{vashishth2019incorporating} applied GCN to incorporate syntactic/semantic information for word embedding training. \citet{cao2019multi} proposed an alignment-oriented knowledge graph embedding for entity alignment. TensorGCN \citep{liu2020tensor} proposed semantic, syntactic, and sequential contextual information. In their framework, multiple aspect graphs are constructed from external resources, and those graph are jointly trained. There are several Multi-aspect, Multi-dimension edge research have been published but none of them are working on the Natural Language Processing field \citep{schlichtkrull2018modeling, khan2019multi, ma2020multi, he2020mv}. 
Recently, graph attention mechanism has been applied in text classification tasks \citep{mei2021graph,liu2021deep,yang2021hgat}. Others focus on using both local and global information \citep{jin2021bite}, multi-modality with text and image information \citep{yang2021multimodal}, enhancing TextGCN with other models \citep{ragesh2021hetegcn} and combining with external knowledge \citep{dai2022graph}.

%$(v, v) \in E$

%$X \in \mathbb{R} ^{n\times m}$

%${x_v \in \mathbb{R}^m}$

%$D_{ii} = \Sigma_j A_{ij}$

%\begin{gather}
% L^{(1)} = \rho(\Tilde{A}XW_0)/
% \end{gather}

% $\Tilde{A} = D^{-\frac{1}{2}} A D^{-\frac{1}{2}}$

% $W_0 \in \mathbb{R}^{m \times k}$

% $\rho(x) = max(0, x)$

% $L^{(0)} = X$

\section{ME-GCN}
\begin{figure}[t]    
         \centering
         \includegraphics[width=1.0\linewidth]{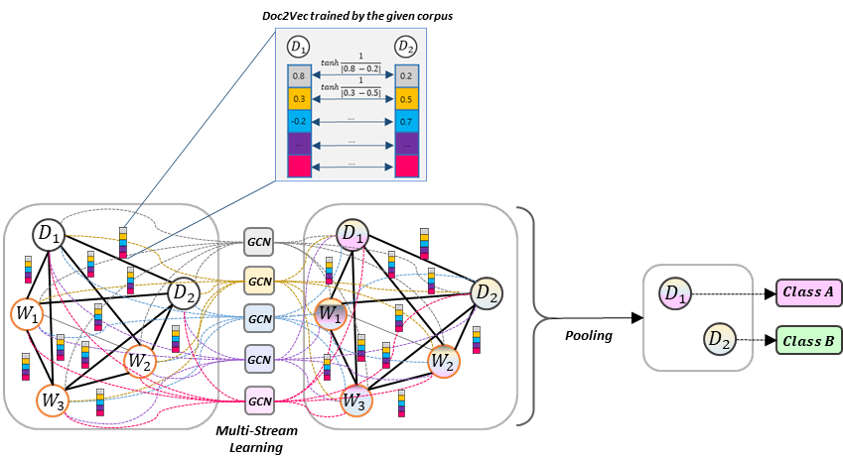}
         \caption{ME-GCN model architecture}
         \label{fig:model}
\end{figure}

We propose the Multi-dimensional Edge-enhanced Graph Convolutional Networks (ME-GCN) for semi-supervised text classification. Note that all graph components are only based on the given text corpus without using any external resources. We utilize the GCN as a base component, due to its simplicity and effectiveness. We first give a brief overview of GCN and introduce details of how to construct our corpus-based textual graph from a given text corpus. Finally, we present ME-GCN learning model. Figure \ref{fig:model} shows the overall architecture of ME-GCN.

\textbf{GCN Graph} A GCN \citep{KipfW17} is a generalised version of the convolutional neural networks for semi-supervised learning that operates directly on the graph-structured data and induces embedding vectors of nodes based on properties of their neighbourhoods. Consider a graph $G = (V, E, A)$, where $V (|V | = N)$ is the set of graph nodes, $E$ is the set of graph edges, and $A \in R^{N \times N}$ is the graph adjacency matrix. 

\subsection{Textual Graph Construction}
We first describe how to construct a textual graph that contains word/document node representation and multi-dimensional edge features for a whole text corpus. We apply a straightforward textual construction approach that treats words and documents as nodes in the graph. Unlike \citep{yao2019graph}, we have three types of edges, namely word-document edge, word-word edge, and document-document edge with the aim to investigate all possible relations between nodes. Formally, we define a ME-GCN graph $G_{ME} = (V, E^{(t)}, ME^{(t)})$, where $t$ denotes the $t^{th}$ dimensional edge, $V(|V|=N)$ is the set of graph nodes of word/document, $E^{(t)}$ are the set of graph edges, which can be one of the three types, and $ME^{(t)}$ is the set of adjacency matrix at the $t^{th}$ dimension. The details of node and edge features construction are presented as follows.

\subsubsection{Textual Node Construction}
From an entire textual corpus, we construct word and document nodes in a graph so that the global word and document distance can be explicitly modeled and graph convolution can be easily adapted. ME-GCN considers the word and document nodes as components for preserving rich information and representing the global structure of a whole corpus, which can fully support for the successful semi-supervised text classification. With this in mind, ME-GCN trains word/node feature by using a Word2Vec \citep{Mikolov2013EfficientEO} for word nodes, and a Doc2Vec \citep{le2014distributed} for document nodes. For instance, Word2Vec takes as its input a whole corpus of words, and the trained word vectors are positioned in a vector space such that words that share common contexts in the corpus are located in close proximity to one another in the space. This is well-aligned with the role of graph neural networks, representing the global structure of the corpus, and preserving rich semantic information of the corpus. Most importantly, those word/document embeddings are distributed representations of text in an $T$-dimensional space so the distance between words and documents can be represented as a multi-dimensional vector. Formally, the word/document node features in ME-GCN are initialised as follows. Note that the negative sampling is applied to reduce the training time.

\textbf{Word Node Construction}
We train the Word2Vec CBOW \citep{Mikolov2013EfficientEO} using context words to predict the centre word. Assume we have a given text corpus consisting of $K$ documents and $U$ unique words. The input is a set of context words $X_{ik}$ in document $k \in K$ encoded as one-hot vector of size $U$. Then the hidden layer $H$ and output layer $Output$ are formulated in equation (1) and (2), in which $W_{U\times T}$ and $W'_{T\times U}$ are two projection matrix. After training, we extract the $U$ vectors of dimension $T$ from the updated matrix $W_{U\times T}$  representing the corresponding $U$ unique words in the whole corpus.
\begin{gather}
H = \sum_{i=1}^{C}X_{ik}W_{U\times T}\\
Output = HW'_{T\times U}
\end{gather}
\textbf{Document Node Construction}
Doc2Vec CBOW \citep{le2014distributed} is essentially the same as Word2Vec. In Doc2Vec, we feed the context words $X_{ik}$ together with the current document $k$ to the model, which is also encoded as one-hot vector based on the document id, and the vector size becomes $\hat{U}=U+K$. We have the projection matrix $W_{\hat{U}\times T}$ containing $U+K$ vectors. After training, those $K$ vectors in the updated $W_{T\times\hat{U}}$ are used for representing the corresponding $K$ document.
\begin{gather}
    H = D_kW_{\hat{U}_\times T}+\sum_{i=1}^{C}X_{ik}W_{\hat{U}\times T}\\
    Output = HW'_{T\times\hat{U}}
\end{gather}

\subsubsection{Multi-dimensional Edge Construction}
In this section, we describe how to construct a multi-dimensional edge feature in a graph. A traditional textual graph edge \citep{yao2019graph} was based on word occurrence in documents (document-word edges), and word co-occurrence in the whole corpus (word-word edges), however, the occurrence information is not enough to extract how close two pieces of text are in both surface proximity and meaning. According to \citet{Mikolov2013EfficientEO, kusner2015word}, the distance between word/document embeddings learn semantically meaningful representations for words from local co-occurrences in sentences and each dimension of word2vec and doc2vec represents the same aspect of word/document representations. Inspired by this, we utilise the distance between each dimension of word/document embeddings to preserve the rich semantic information captured by edges, which are also presented as multi-dimensional vectors. To represent all possible edge types, we propose three types of edges: word-word edges, document-document edges, and word-document edges. Our goal is to incorporate the semantic similarity between individual node pairs (each unique word and document) into multi-dimensional edge features. One such measure for word/document node similarity is provided by their Euclidean distance in the Word2Vec or Doc2Vec embedding space. We separately use each dimension space in the node feature (Word2Vec/Dec2Vec) for representing each of the dimension in the multi-dimensional node edge. Thus, we have $T$ dimensional edges between nodes of $T$ dimensional features and each $t \in \{1,2,...,T\}$ is represented by one dimensional Euclidean distance calculation in the $t^{th}$ dimensional space. This edge calculation method is applied to word-word and doc-doc edges. 

\textbf{Word-Word Edge Feature}
We draw on the learned semantics in each feature dimension of the word embedding of size $T$ to calculate the edge weight for each dimension. Concretely, the $T$-dimensional word-word edge $E_{w_i,w_j}^{(t)}, t \in \{1,2,...,T\}$ between word $i$ and word $j$ is formulated as in equation (5), in which $W_i^{(t)}$ and $W_j^{(t)}$ represents the feature value at the dimension $t$ of the word embedding $W_i$ for word i and $W_j$ for word j respectively. The denominator calculates the distance of the two words regarding dimension $t$ and $tanh(^{-1})$ is used for normalization. 
\begin{gather}
E_{w_i,w_j}^{(t)} = tanh{1\over|W_i^{(t)} - W_j^{(t)}|}
\end{gather}
\textbf{Doc-Doc Edge Feature} 
The document-document edge is constructed in a way similar to the word-word edge. As is shown in equation (6), the $T$-dimensional document-document edge $E_{d_i,d_j}^{(t)}$ is calculated based on the normalized Euclidean distance between the values $D_i^{(t)}$ and $D_j^{(t)}$ at each dimension $t$ of the features for document $i$ and $j$. To relieve over-smoothing issue, we only consider edges between two documents having over $u$ overlapping words. 
\begin{gather}
E_{d_i,d_j}^{(t)} = tanh{1\over|D_i^{(t)} - D_j^{(t)}|} \text{    if  } W_{d_i\cap d_j}\geq u
\end{gather}
\textbf{Word-Doc Edge Feature} 
We use the same calculation method for a single-dimension word-document edge as in TextGCN while repeating it for each dimension $t$. Thus, the $T$-dimensional word-document edge $E_{w_i,d_j}^{(t)}$ is simply represented as the TF-IDF value of word $i$ and document $j$. This is repeated for each dimension $t$, as is formulated in equation (7). We also found using TF-IDF weight is better than using term frequency only.
\begin{gather}
E_{w_i,d_j}^{(t)} = \text{TF-IDF}_{w_i,d_j}
\end{gather}

% \begin{equation}
%   \small 
%     ME_{ij}^{(t)} = \begin{cases}
%     E_{w_i,w_j}^{(t)}&w_i, w_j\text{ are words}\\
%     E_{d_i,d_j}^{(t)}&d_i,d_j\text{ are docs, } W_{d_i\cap d_j}\geq u\\
%     E_{w_i,d_j}^{(t)}&w_i\text{ is word}, d_j \text{ is doc}\\
%     1&i=j\\
%     0&\text{otherwise}
%     \end{cases}
% \end{equation}

% Formally, the multi-dimensional edge weights between node $i$ and $j$ is defined as in equation (8). 
% We noted that the threshold $u$ for the doc-doc edges is not compulsory but efficient for the better computation. The detailed threshold is described in Section \ref{sec:hp}.

%the formulation of our final adjacency matrix $ME_{ij}^{(t)}$ for the the graph $G_{ME}$. The multi-dimensional edge for node $i$ and $j$ can be $E_{w_i,w_j}^{(t)}$ when $i$ and $j$ are both word nodes; or $E_{d_i,d_j}^{(t)}$ when $i$ and $j$ are both document nodes and the number of shared unique words by the two documents $W_{d_i\cap d_j}$ is no less than a threshold $u$; or $E_{w_i,d_j}^{(t)}$ the two nodes are word-document pair; or the edge is set to 1 for self-connection ($i=j$); otherwise no edge will be connected between the two nodes.  

\subsection{ME-GCN Learning}
%Take a brief revisit to traditional GCN learning. Consider a graph $G = (V, E, A)$, in which $V$ is the set of graph nodes ($|V|=N$), $E$ is the set of graph edges, and $A \in R^{N \times N}$ is the graph adjacency matrix. 
After constructing the multi-dimensional edge enhanced text graph, we focus on applying effective learning framework to perform GCN on the textual graph with multi-dimensional edge features. 

The traditional GCN learning takes into the initial input matrix $H^{(0)} \in R^{N \times d_0}$ containing $N$ node features of size $d_0$. Then the propagation through layers is made based on the rule in equation (9), which takes into consideration both node features and the graph structure in terms of connected edges.  
\begin{gather}
H^{(l+1)} = f(H^{(l)}, A) = \sigma (\hat{A}H^{(l)}W^{(l)})
\end{gather}

\noindent The $l$ and $(l+1)$ represents the two subsequent layers, $\hat{A}=\tilde{D}^{-\frac{1}{2}}\tilde{A}\tilde{D}^{-\frac{1}{2}}$ is the normalized symmetric adjacency matrix $\tilde{A} = A+I$ ($I$ is an identity matrix for including self-connection), $\tilde{D}$ is the diagonal node degree matrix with $\tilde{D}(i,i)=\Sigma_j\tilde{A}(i,j)$, and $W^{(l)} \in R^{d_l\times d_{l+1}}$ is a layer-specific trainable weight matrix for $l$th layer. $d_l$ and $d_{l+1}$ indicates the node feature dimension for $l$th layer and $(l+1)$th respectively. $\sigma$ denotes a non-linear activation function for each layer such as Leaky ReLu/ReLU except for the output layer where softmax is normally used for the classification. 

Our goal is to represent the node representation by aggregating neighbour information with each edge features in a multi-stream manner. Hence, we generalize the traditional GCN learning approach to perform multi-stream(MS) learning for the multi-dimensional edge enhanced graph. The overall MS learning procedure is in equation (10), for each node feature in $H^{(l)} \in R^{N \times d_l}$, we apply the multi-stream GCN learning $f_{MS}$ that formulates $t$ streams of traditional GCN learning in equation (9) through the $t$ dimensions of the connected edge, resulting in the multi-stream hidden feature $H_t^{(l+1)} \in R^{N \times d_{ms}^{(l+1)}}$ at $(l+1)$th layer. Here $t \in \{1,2,...,T\}$ and $d_{ms}^{(l+1)}$ is the multi-stream feature size for each edge dimension at this layer. Then a multi-stream aggregation function $\phi_{MS}$ is applied over the $t$ streams, producing the feature matrix $H^{(l+1)} \in R^{N \times d_{(l+1)}}$ that contains the aggregated feature for each node in $N$. Here we use $concatenation$ function as $\phi_{MS}$ for the hidden layer in the multi-stream aggregation, leading us to have  $d_{l+1}=t*d_{ms}^{(l+1)}$. Specifically, for the output layer, $pooling$ method is used instead and the details are provided in later paragraph. Accordingly, the updated propagation rule is provided in equation (11). Unlike the original GCN propagation in equation (9), we have $T$ streams of GCN learning in each layer, sharing the same input $H^{(l)}$ and propagating based on the $T$ adjacency matrices $ME^{(t)}$, which involves a set of layer and stream specific trainable weight matrices denoted as $W^{(l)(t)}$. We also tried the shared-stream learning that shares the trainable weight matrices across each stream but found that separate stream-specific trainable weight matrices have better performance. The comparison of the two learning mechanisms is provided in Appendix Section \ref{sec:pooling_experiment}. 
\begin{gather}
H^{(l)} \xrightarrow{f_{MS}} H_{t}^{(l+1)} \xrightarrow{\phi_{MS}} H^{(l+1)}
\end{gather}
\begin{gather}
H^{(l+1)} = \phi_{MS}(f_{MS}(H^{(l)}, ME^{(t)})
)  \\
= \phi_{MS}( \sigma (\hat{ME}^{(t)}H^{(l)}W^{(l)(t)})) \notag
\end{gather}

% \subsubsection{Pooling}\label{sec:pooling}
Unlike the hidden layers where we use $concatenation$ to aggregate the node features over each stream to continue propagation to next layer, we instead apply the $pooling$ at the output layer to further synthesize the multi-stream features of each node to do the final classification. Equation (12) formulizes $max$ $pooling$, in which $H_t^{(l_O)} \in R^{N \times d^{l_O}_{ms}}, t \in \{1,2,...,T\}$ denotes the $T$ streams of node features for $N$ nodes at the output layer $l_O$, and here $d^{l_O}_{ms}$ is the node feature dimension that equals to the classification label number $C$. Through $max$ $pooling$, we select the best valued features over the $T$ streams for each node in $N$ before the final classification. We also tried other $pooling$ and provide the comparison in Appendix Section \ref{sec:pooling_experiment}.
\begin{gather}
pooling_{max}= \smash{\displaystyle\max_{1 \leq t \leq T}}(H_{t}^{(l_{O})})
\end{gather}
%In practise, we tried three different types of pooling methods, which are $max$ $pooling$, $average$ $pooling$, $min$ $pooling$. We provide the equation (12) for $max$ $pooling$, in which $H_t^{(l_O)}$ denotes the $T$ streams of node features at the output layer $l_O$, and $C$ is the node feature dimension that equals the classification label number. Through $max$ $pooling$, we pick out the max valued features over the $T$ streams for each node in $N$. Similar to $max$ $pooling$, $average$ $pooling$ will make a balance across the stream and $min$ $pooling$ will focus on the smallest valued features by replacing the $max$ operation to $average$ and $min$ respectively. Our experiment shows that $min$ $pooling$ always got the worse performance while $max$ $pooling$ worked better the the other two $pooling$ methods over most of the experiment datasets. We provide the comparison among these three $pooling$ methods in Section xxx.
\section{Evaluation Setup}

We evaluate our ME-GCN on text classification in semi-supervised settings, and examine the effectiveness of corpus-based multi-dimensional edge features.

\textbf{Baselines:} We compare ME-GCN with state-of-the-art semi-supervised text classification models, which do not use any external resources. Additionally, we also include four baseline models, which use pretrained embedding or language model: CNN-Pretrained, LSTM-Pretrained, BERT, and TMix. \textbf{1)TF-IDF+LR, 2)TF-IDF+SVM}: Term frequency inverse document frequency for feature engineering with Logistic Regression or SVM with rbf kernel. \textbf{3)CNN-Rand, 4)-Pretrained}: Text-CNN \citep{kim2014convolutional} is used as the classifier. Both CNN-Rand using random initialized word embedding and CNN-Pretrained using pretrained word embedding are evaluated. We used English Glove-pretrained \citep{pennington2014glove} and Chinese Word Vectors \citep{li2018analogical} for Chinese dataset-zh. \textbf{5)LSTM-Rand, 6)-Pretrained}: We apply the same set-up as the CNN, but with Long Short-Term Memory (LSTM). \textbf{7)TextGCN}: We follow the same hyperparameters of the TextGCN \citep{yao2019graph}. \textbf{8)BERT}: We use huggingface\citep{wolf-etal-2020-transformers} BERT\textsubscript{BASE} \citep{devlin2018bert}  in our experiments (`bert-base-chinese' model is used for Chinese). \textbf{9)TMix}: TMix\citep{chen2020mixtext} generates new training text data by interpolating over labelled text encoded using BERT hidden representation and train on the generated text data for text classification. We use the default setting provided.

\begin{table*}[!t]
\centering
\begin{adjustbox}{width=1\linewidth}
\centering
\begin{tabular}{l|c|cccccccc}
\hline
 \multicolumn{1}{c|}{\textbf{Methods}} & \textbf{Pretrained} & \textbf{20NG}   & \textbf{R8}  & \textbf{R52}  & \textbf{Ohsumed} & \textbf{MR} & \textbf{Agnews} & \textbf{Twit nltk} & \textbf{Waimai(zh)}  \\ \hline
TFIDF + SVM & \xmark & 0.2529 & 0.7246 & 0.5932 & 0.1589 & 0.5884 & 0.4241 & 0.5737 & 0.7521\\
TFIDF + LR & \xmark & \underline{0.2633} & 0.7249 & 0.6332 & 0.1798 & 0.5871 & 0.5370 & 0.5791 & 0.7381\\
CNN - Rand & \xmark & 0.0768 & 0.7219 & 0.6325 & 0.1889 & 0.5641 & 0.3825 & 0.5822 & 0.7784\\
CNN - Pretrained & \cmark & 0.2380 & 0.7428 & \underline{0.6896} & \underline{0.2458} & 0.6005 & 0.6636 & 0.6088 & 0.7926\\
LSTM - Rand & \xmark & 0.0545 & 0.6788 & 0.4253 & 0.1319 & 0.5442 & 0.3444 & 0.5458 & 0.6458\\
LSTM - Pretrained & \cmark & 0.0593 & 0.6919 & 0.5285 & 0.0948 & 0.5933 & 0.5815 & 0.6098 & 0.6663\\
TextGCN & \xmark & 0.1188 & \underline{0.8628} & 0.4847 & 0.1612 & 0.6222 & 0.7420 & \underline{0.7806} & 0.8065\\
BERT  & \cmark & 0.1347 & 0.5148 & 0.6291 & 0.1464 & \textbf{0.7666} & 0.7261 &  0.7024 & \underline{0.8248}\\

TMix & \cmark & 0.2286 & 0.7322 & 0.6195 & 0.1721 & 0.6267 & \underline{0.8025} & 0.6111 & 0.6376\\\hline
Our ME-GCN & \xmark & \textbf{0.2861} & \textbf{0.8679} & \textbf{0.7828} & \textbf{0.2740} & \underline{0.6811} & \textbf{0.8043} & \textbf{0.8232} & \textbf{0.8393}\\\hline
\end{tabular}
\end{adjustbox}
\caption{Test accuracy comparison. The bottom row shows the best test accuracy from our proposed model using either max pooling or average pooling. The comparison of our model performance for each dataset using the three pooling methods is provided in Table \ref{poolingtable}. The second best is \underline{underlined}.}
\label{baselinetable}
\end{table*}

\textbf{Dataset: } We evaluated our experiments on five widely used text classification benchmark datasets \citep{yao2019graph}, 20NG, R8, R52, MR and Ohsumed, and three additional semi-supervised text classification datasets \citep{linmei2019heterogeneous}, Agnews, Twitter nltk and Waimai. All the data is split based on the extreme low resource text classification enviornment- 1\% training and 99\% test set. The summary statistics of the datasets can be found in Appendix Table \ref{datasettable}. For the data sample selection, we randomly select them but the class distribution is followed by the original datasets. \textbf{1)20NG} is a 20-class news classification dataset and we select 3,000 samples from the original dataset. \textbf{2)R8}, \textbf{3)R52} are from Reuters which is a topic classification dataset with 8 classes and 52 classes. 3,000 samples from each dataset are selected. \textbf{4)MR}\citep{pang2005seeing} is a binary classification dataset about movie comments and we use all samples from the dataset. \textbf{5)Ohsumed} is a medical dataset with 23 classes, and we select 3,000 samples from the original dataset. \textbf{6)Agnews}\citep{zhang2015character} is a 4-class news classification dataset and 6,000 samples are selected. \textbf{7)Twitter nltk} is a binary classification sentiment analysis from Twitter, we sampled 1,500 positive and 1,500 test samples from the original dataset. \textbf{8)Waimai} is a binary sentiment analysis dataset about food delivery service comments from a Chinese online food ordering platform. The dataset is in Chinese and pre-tokenized.

%the default number of streams is 25, which is the dimension of corpus-trained word and document embeddings and also the edge feature dimension. Different number of streams are tested and discussed in \ref{edge_feature_section}. The default value of document threshold $u$ for document-document edge construction is 5. For both Word2vec/Doc2vec training, words occurring fewer than 5 times have been excluded before training process. 

%In the training process, followed by \citet{liu2020tensor}, we use dropout rate as 0.5 and learning rate as 0.002 with Adam optimizer. The number of epochs is 2000 and 10\% of the training set is used as the validation set for early stopping when there is no decreasing in validation set's loss for 100 consecutive epochs.

\section{Results Analysis}

\subsection{Performance Evaluation}
Table \ref{baselinetable} presents a comprehensive performance experiment, conducted on the benchmark datasets. The most bottom row shows the accuracy from our best models using either max or average pooling.

Overall, our proposed model significantly outperforms the baseline models on all eight datasets, demonstrating the effectiveness of our ME-GCN on semi-supervised text classification for various length of text. With in-depth analysis, CNN/LSTM-Rand is quite low in performance on several datasets but increases significantly when using pretrained embeddings. While TextGCN achieves better accuracy than above baselines on most datasets, the performance is all lower than ME-GCN. This shows the efficiency of preserving rich information using multi-dimensional edge features. The merit of pre-training stands out with BERT and TMix, producing better accuracy than the baseline TextGCN on most datasets. Especially, BERT achieves the best and second best performance on MR and Waimai, which are short-text sentiment analysis datasets. This would be because of the two aspects of sentiment classification: (1) compared to topic-specific text classification, sentiment analysis task may benefit from the pretrained general semantics learned from a large external text; (2) word order matters for sentiment analysis, which could be missing in GNNs. Nevertheless, our ME-GCN, with no external resources, still outperforms those pertrained models in seven datasets, illustrating the potential superiority of self-exploration on the corpus via multi-dimensional edge graph in comparison of pretraining on large external resource. 

%Overall, our proposed model significantly outperforms the baseline models on six of the datasets, demonstrating the effectiveness of our ME-GCN on semi-supervised text classification for various length of text datasets. With in-depth analysis, CNN/LSTM-Rand is quite low but increases significantly when those models with pretrained embeddings. While TextGCN achieves better accuracy than above baselines in most datasets, the performance is all lower than ME-GCN. This shows the efficiency of preserving rich information using multi-dimensional edge features. 

%However, the merit of pre-training stands out again with BERT for short text and sentiment analysis datasets, such as MR, Twit nltk and Waimai, further outperforming TextGCN. This might be attributed to the nature of sentiment classification in terms of two aspects: (1) compared to topic-specific text classification (e.g. news classification), sentiment analysis task may better benefit from the pretrained general semantics learned from a large external text; (2) word order matters for sentiment analysis, which could be missing in GNNs. However, in Waimai(in Chinese), our ME-GCN is slightly higher than the BERT. We can say the limitation of GNN may be less in Chinese. 

\begin{figure}[t]    
     \begin{subfigure}{0.24\linewidth}
         \includegraphics[width=\linewidth]{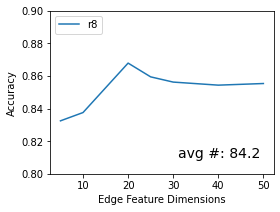}
         \caption{R8}
         \label{fig:r8_dim}
     \end{subfigure}
          \begin{subfigure}{0.24\linewidth}
         \includegraphics[width=\linewidth]{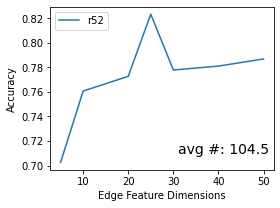}
         \caption{R52}
         \label{fig:r52_dim}
     \end{subfigure}
          \begin{subfigure}{0.24\linewidth}
         \includegraphics[width=\linewidth]{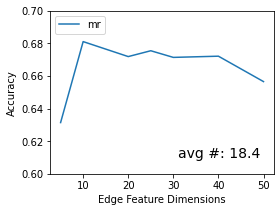}
         \caption{MR}
         \label{fig:mr_dim}
     \end{subfigure}
          \begin{subfigure}{0.24\linewidth}
         \includegraphics[width=\linewidth]{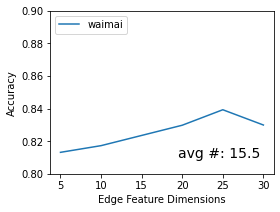}
         \caption{Waimai(zh)}
         \label{fig:waimai_dim}
     \end{subfigure}
     \caption{Test accuracy by varying edge feature dimensions. The bottom right corner shows the average number of words per document in each corpus.}
     \label{fig:dim_result}
\end{figure}

\subsection{Impact of Edge Feature Dimension}\label{edge_feature_section}
To evaluate the effect of the dimension size of the edge features, we tested ME-GCN with different dimensions. Figure \ref{fig:dim_result} shows the test accuracy of our ME-GCN model on the four dataset, including R8, R52, MR, Waimai(zh). The bottom right corner for each subgraph includes the average number of the words per document.
We noted that the test accuracy is related to the average number of words per document in the corpus. For instance, for ‘MR’ (avg \#: 18.4), test accuracy first increases with the increase of the size of edge feature dimensions, reaching the highest value at 10; it falls when its dimension is higher than 15. However, for R8 and R52 (avg \#: 84.2 and 104.5), got the highest value at 20 or 25. This is consistent with the intuition that the average number of words per document in the corpus should align with the dimension size of the edge features in ME-GCN. The trend is different in waimai dataset as it is Chinese, this is because different languages would have different nature of choosing the efficient edge feature dimension.

\begin{figure}[t]    
    \begin{subfigure}{0.24\linewidth}
         \includegraphics[width=\linewidth]{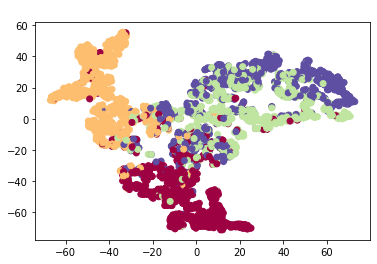}
         \caption{Dim = 5, second layer}
         \label{fig:agnews_5_2}
    \end{subfigure}
    \begin{subfigure}{0.24\linewidth}
         \includegraphics[width=\linewidth]{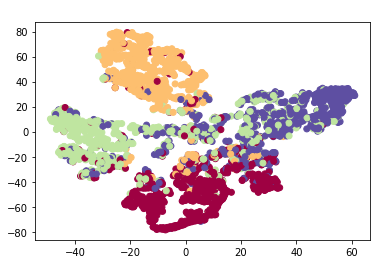}
         \caption{Dim = 25, second layer}
         \label{fig:agnews_25_2}
    \end{subfigure}
     \begin{subfigure}{0.24\linewidth}
         \includegraphics[width=\linewidth]{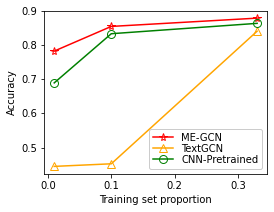}
         \caption{R52}
         \label{fig:r52_pro}
    \end{subfigure}
    \begin{subfigure}{0.24\linewidth}
         \includegraphics[width=\linewidth]{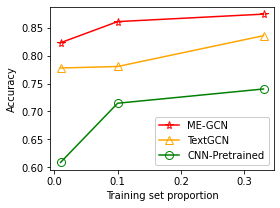}
         \caption{Twitter nltk}
         \label{fig:twitter_pro}
    \end{subfigure}
     \caption{(a)(b)t-SNE visualisation of test set document embeddings in AgNews (4 classes). The (a) and (b) show second layer document embeddings learned by 5 and 25 dimensional node and edge features respectively.(c)(d)Test accuracy comparison with different number of labelled documents.}
     \label{fig:tsne}
\end{figure}

Moreover, in order to analyse the impact of the edge feature dimension, we present an illustrative visualisation of the document embeddings learned by ME-GCN. We use the t-SNE tool \citep{van2008visualizing} in order to visualise the learned document embeddings. Figure \ref{fig:agnews_5_2} and Figure \ref{fig:agnews_25_2} shows the visualisation of test set document embeddings in AgNews learned by ME-GCN (second layer) 5 and 25 dimensional node and edge features. The AgNews has 4 classes and the average number of words per document is 35.2. Instead of dim=5, having dim=25 as edge features would better to separate them into four classes. 

\subsection{Impact of Ratio of Labelled Docs}
We choose 3 representative methods with the best performance from Table \ref{baselinetable}: CNN-Pretrained, TextGCN and our ME-GCN, in order to study the impact of the number of labelled documents. Particularly, we vary the ratio of labelled documents and compare their performance on the two datasets, Twitter nltk and R52, that have the smallest number and largest number of classes. Figure \ref{fig:r52_pro} and Figure \ref{fig:twitter_pro} reports test accuracies with 1\%, 10\%, and 33\% of the R52 and Twitter nltk training set. We note that our ME-GCN outperforms all other methods consistently. For instance, ME-GCN achieves a test accuracy of 0.8232 on Twitter nltk with only 1\% training documents and a test accuracy of 0.8552 on R52 with only 10\% training documents which are higher than other models with even the 33\% training documents. It demonstrates that our method can more effectively take advantage of the limited labelled data for text classification.

\begin{table*}[t]
\centering
\begin{adjustbox}{width=1\linewidth}
\centering
\begin{tabular}{c|cccccccc}
\hline
\textbf{Word Embedding}  & \textbf{20NG}   & \textbf{R8}  & \textbf{R52}  & \textbf{Ohsumed} & \textbf{MR} & \textbf{Agnews} & \textbf{Twit nltk} & \textbf{Waimai(zh)}  \\ \hline
Word2Vec & \textbf{0.2861} & \textbf{0.8679} & 0.7828 & 0.2740 & 0.6811 & \textbf{0.8043} & 0.8232 & \textbf{0.8393}\\
fastText & 0.2510 & 0.8394 & 0.7783 & 0.2550 & 0.6727 & 0.7812 & 0.8333 & 0.8191\\
GloVe & 0.2526 & 0.8247 & \textbf{0.7835} & \textbf{0.2832} & \textbf{0.6895} & 0.7628 & \textbf{0.8341} & 0.8298\\\hline
\end{tabular}
\end{adjustbox}
\caption{Test accuracy comparison of our ME-GCN model with different word embedding techniques to train word node embeddings and word-word multi-dimensional edge features.}
\label{embeddingtable}
\end{table*}

\subsection{Comparison of Embedding Variants}
ME-GCN apply a Word2Vec CBOW in order to train the word node embedding and the related multi-dimensional edge feature. We compare our model with three different word embedding techniques, Word2Vec, fastText, and Glove in Table \ref{embeddingtable}. We noted that using Word2Vec and Glove, word-based models, is comparatively higher than applying the fastText, a character n-gram-based model. This would be affected because the node and edge of ME-GCN are based on words, not characters.

\begin{table*}[!t]
\centering
\begin{adjustbox}{width=1\linewidth}
\centering
\begin{tabular}{c|cccccccc}
\hline
\textbf{Pooling Method}  & \textbf{20NG}   & \textbf{R8}  & \textbf{R52}  & \textbf{Ohsumed} & \textbf{MR} & \textbf{Agnews} & \textbf{Twit nltk} & \textbf{Waimai(zh)}  \\ \hline
Max Pooling & 0.2775 & 0.8473 & \textbf{0.7828} & 0.2475 & \textbf{0.6811} & \textbf{0.8043} & \textbf{0.8232} & \textbf{0.8393}\\
Avg Pooling & \textbf{0.2861} & \textbf{0.8679} & 0.7675 & \textbf{0.2740} & 0.6658 & 0.7911 & 0.8205 & 0.8303\\
Min Pooling & 0.0424 & 0.2987 & 0.2550 & 0.0294 & 0.5000 & 0.2005 & 0.5000 & 0.6663\\\hline
\end{tabular}
\end{adjustbox}
\caption{Test accuracy of ME-GCN with three different pooling methods, max, average, and min pooling}
\label{poolingtable}
\end{table*}

\begin{table*}[!t]
\centering
\begin{adjustbox}{width=1\linewidth}
\centering
\begin{tabular}{c|cccccccc}
\hline
\textbf{Learning Methods}  & \textbf{20NG}   & \textbf{R8}  & \textbf{R52}  & \textbf{Ohsumed} & \textbf{MR} & \textbf{Agnews} & \textbf{Twit nltk} & \textbf{Waimai(zh)} \\ \hline
Separated Learning & \textbf{0.2861} & \textbf{0.8679} & \textbf{0.7828} & \textbf{0.2740} & \textbf{0.6811} & \textbf{0.8043} & \textbf{0.8232} & \textbf{0.8393} \\
Shared Learning & 0.1582 & 0.8016 & 0.6554 & 0.2635 & 0.6575 & 0.6993 & 0.7037 & 0.8137\\\hline
\end{tabular}
\end{adjustbox}
\caption{Test accuracy of ME-GCN with two multi-stream learning methods, shared and separated learners.}
\label{learningtable}
\end{table*}

\subsection{Learning and Pooling Variant Testing}\label{sec:pooling_experiment}
We compare ME-GCN with three different pooling approaches (max, average, and min pooling) and the result is shown in Table \ref{poolingtable}. Most datasets produce better results when using max pooling, and the result with max and average pooling outperforms that with min pooling. This is very obvious because the min pooling captures the minimum value of each graph component. 

We also compare two multi-stream graph learning methods, including separated and shared stream learning to examine the effectiveness of ME-GCN learning with multi-dimensional edge features. Table \ref{learningtable} presents that the separated stream learners significantly outperforms the shared learners. This shows it is much efficient to learn each dimensional stream with an individual learning unit and initially understand the local structure, instead of learning all global structures at once.

\section{Conclusion}
We introduced ME-GCN (Multi-dimensional Edge-enhanced Graph Convolutional Networks) for semi-supervised text classification, which takes full advantage of both limited labelled and large unlabelled data by rich node and edge information propagation. We propose corpus-trained multi-dimensional edge features to efficiently handle the distance/closeness between words and documents as multi-dimensional edge features, and all graph components are based on the given text corpus only. ME-GCN demonstrates promising results by outperforming numerous state-of-the-arts on eight semi-supervised text classification datasets consistently. In the future, it would be interesting to make this multi-aspect graph under inductive learning.

\bibliography{iclr2022_conference}
\bibliographystyle{iclr2022_conference}

\appendix

% \begin{table*}[t]
% \begin{adjustbox}{width=1\linewidth}
% \begin{tabular}{c|c|c|c|c|c|c|c|c|c}
% \hline
% \multicolumn{2}{c|}{}  & \textbf{20NG}   & \textbf{R8}  & \textbf{R52}  & \textbf{Ohsumed} & \textbf{MR} & \textbf{Agnews} & \textbf{Twit nltk} & \textbf{Waimai(zh)}  \\ \hline
% \multirow{2}{*}{Word Node} & \# Parameters & 304,750 & 217,650 & 230,950 & 432,950 & 225,050 & 268,000 & 31,700 & 548,950\\\cline{2-10}
% & Running Time(s) & 118 & 25 & 76 & 140 & 71 & 83 & 20 & 74\\\hline
% \multirow{2}{*}{Doc Node} & \# Parameters & 454,750 & 367,650 & 380,950 & 582,950 & 758,150 & 568,000 & 181,700 & 1,148,300\\\cline{2-10}
% & Running Time(s) & 104 & 35 & 118 & 272 & 270 & 140 & 29 & 312\\\hline
% \multirow{2}{*}{Model Learning} & \# Parameters & 328,125 & 140,625 & 828,125 & 375,000 & 46,875 & 78,125 & 46,875 & 46,875\\\cline{2-10}
% & Running Time(s) & 198 & 16 & 164 & 286 & 120 & 612 & 14 & 610\\\hline
% \multirow{2}{*}{Total} & \# Parameters & 1,087,625 & 725,925 & 1,440,025 & 1,390,900 & 1,030,075 & 914,125 & 260,275 & 1,744,125\\\cline{2-10}
% & Running Time(s) & 420 & 76 & 358 & 698 & 461 & 835 & 63 & 996\\\hline
% \end{tabular}
% \end{adjustbox}
% % \setlength{\belowcaptionskip}{-10pt}
% \caption{Number of Parameters and Running time for each dataset}
% \label{tab:detailtable}
% \end{table*}

\section{Settings}
\subsection{Hyperparameter Setting}\label{sec:hp}
All documents are tokenized using NLTK tokenizer\citep{bird2009natural}, and words occurring no more than 5 times have been excluded. Both word2vec and Dec2vec are trained on the corpus we get using $gensim$ package with $window\_size=5$ and $iter=200$.  The initial feature dimension for node and document is set to $d_0=25$, which is same to the multi-dimension number for edge features and multi-stream number $T$ in ME-GCN learning. Different multi-stream numbers are tested and discussed in Section \ref{edge_feature_section}. The threshold $u=5$ is used for document-document edge construction. We use two-layers of multi-stream GCN learning with $d^{l_1}_{ms} =25$ (thus $d^{l_1} =625$) for the first multi-stream GCN layer and $d^{l_O}_{ms}=C$(no. of label in the datasets) for the output layer. In the training process, following \citet{liu2020tensor}, we use dropout rate as 0.5 and learning rate as 0.002 with Adam optimizer. The number of epochs is 2000 and 10\% of the training set is used as the validation set for early stopping when there is no decreasing in validation set's loss for 100 consecutive epochs.

\begin{table}[h]
\centering
\begin{adjustbox}{width=0.65\linewidth}
\begin{tabular}{c|ccccc}
\hline
\textbf{Datasets}  & \textbf{\# Doc}   & \textbf{\# Words}  & \textbf{\# Node}  & \textbf{\# Class}  &
\textbf{Avg. length}\\ \hline
20NG & 3,000 & 6,095 & 9,095 & 20 & 249.4\\ 
R8 & 3,000 & 4,353 & 7,353 & 8 & 84.2\\
R52 & 3,000 & 4,619 & 7,619 & 52 & 104.5\\
Ohsumed & 3,000 & 8,659 & 11,659 & 23 & 132.6\\
MR & 10,662 & 4,501 & 15,163 & 2 & 18.4\\
Agnews & 6,000  & 5,360 & 11,360 & 4 & 35.2\\
Twit nltk & 3,000  & 634 & 3,634 & 2 & 11.5\\
Waimai(zh)& 11,987  & 10,979 & 22,966 & 2 & 15.5\\\hline
\end{tabular}
\end{adjustbox}
\caption{The summary statistics of datasets}
\label{datasettable}
\end{table}

% \section{Model Architecture}
% The overall architecture of ME-GCN is presented in Figure \ref{fig:model}. After the paper get accepted, the architecture figure will be moved to the main content. 

\subsection{Hyperparemeter Search}
For each dataset we use grid search to find the best set of hyperparameters and select the base model based on the average accuracy by running each model for 5 times. The number of stream: 5,10,20,25,30,40,50. The document edge threshold: 3,5,10,15. The pooling method: max pooling, min pooling, average pooling. The number of hyperparameter search trials is 72($=6*4*3$) for each dataset. The best hyperparameters for each dataset and their average accuracy on test set shows in Table \ref{tab:parametertable}. And the trend of validation performance is very similar to the testing performance trend.

\begin{table*}[h]
\centering
\begin{adjustbox}{width=1\linewidth}
\centering
\begin{tabular}{c|c|c|c|c|c|c|c|c}
\hline
  & \textbf{20NG}   & \textbf{R8}  & \textbf{R52}  & \textbf{Ohsumed} & \textbf{MR} & \textbf{Agnews} & \textbf{Twit nltk} & \textbf{Waimai(zh)}  \\ \hline
\# Stream & 30 & 20 & 25 & 30 & 10 & 20 & 25 & 30\\\hline
Document Threshold & 15 & 10 & 15 & 5 & 5 & 5 & 3 & 3\\\hline
Pooling Method & avg & avg & max & avg & max & avg & max & max\\\hline
Accuracy & 0.2861 & 0.8679 & 0.7828 & 0.2740 & 0.6811 & 0.8043 & 0.8232 & 0.8393\\\hline
\end{tabular}
\end{adjustbox}
\caption{Best hyperparameters for each dataset}
\label{tab:parametertable}
\end{table*}

\subsection{Running Details}
All the models are trained by using 16 Intel(R) Core(TM) i9-9900X CPU @ 3.50GHz and NVIDIA Titan RTX 24GB using Pytorch \citep{NEURIPS2019_9015}. 

\begin{figure}[h]    
         \centering
         \includegraphics[width=0.6\linewidth]{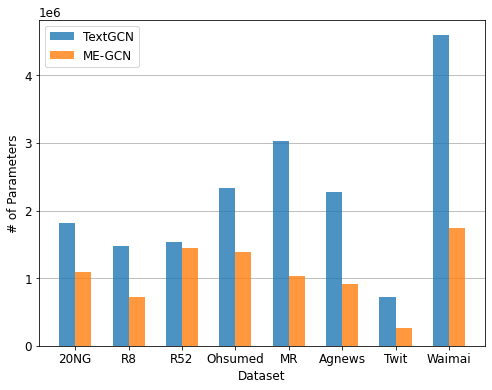}
         \caption{Number of Parameters Comparison}
         \label{fig:para_comp}
\end{figure}

The number of parameters for each part of the model is: Word Node (Word2vec): $2UT$, Document Node (Doc2vec): $2T(U+K)$, ME-GCN Learning: $T^2d_{ms}^{l_1}(1+C)$. The default value of $d_{ms}^{l_1}$ is 25. The number of parameters of TextGCN is $(U+K)*D+D*C$ and the default value of $D$ is 200. Comparison of the number of parameters between TextGCN and our ME-GCN shows in Figure \ref{fig:para_comp}.

\section{Links Related to Datasets and Baseline Models}
The links for Datasets:
\begin{itemize}
    \item \textbf{20NG}: \url{http://qwone.com/~jason/20Newsgroups/}
    \item \textbf{R8}, \textbf{R52}: \url{https://www.cs.umb.edu/~smimarog/textmining/datasets/}
    \item \textbf{MR}: \url{http://www.cs.cornell.edu/people/pabo/movie-review-data/}
    \item \textbf{Ohsumed}: \url{http://disi.unitn.it/moschitti/corpora.htm}
    \item \textbf{Agnews}: \url{http://www.di.unipi.it/~gulli/AG_corpus_of_news_articles}
    \item \textbf{Twitter nltk}: \url{http://nltk.org/howto/twitter.html}
    \item \textbf{Waimai}: \url{https://github.com/SophonPlus/ChineseNlpCorpus/}
\end{itemize}

The links for Baseline Models:
\begin{itemize}
    \item \textbf{TextCNN}: \url{https://github.com/DongjunLee/text-cnn-tensorflow}
    \item \textbf{TextGCN}: \url{https://github.com/yao8839836/text\_gcn}
    \item \textbf{BERT BASE}: \url{https://huggingface.co/bert-base-uncased}
    \item \textbf{Tmix}: \url{https://github.com/GT-SALT/MixText}
    \item \textbf{Chinese BERT}: \url{https://huggingface.co/bert-base-chinese}
    \item \textbf{GloVe-pretrained}: \url{https://nlp.stanford.edu/projects/glove/}
    \item \textbf{Chinese Word Vectors}: \url{https://github.com/Embedding/Chinese-Word-Vectors}
\end{itemize}

The tokenizer used:
\begin{itemize}
    \item \textbf{English Tokenizer - NLTK}:
    \url{https://www.nltk.org/api/nltk.tokenize.html}
    \item \textbf{Chinese Tokenizer - Jieba}: \url{https://github.com/fxsjy/jieba}
\end{itemize}

\end{document}